\documentclass[runningheads]{llncs}
\usepackage{amsmath}
\usepackage{graphicx}
\usepackage{tabularx}
\usepackage[table]{xcolor}
\usepackage{colortbl}
\usepackage{float}
\usepackage{caption}
\usepackage{subcaption}
\usepackage{makecell}
\usepackage{url}
\usepackage{nth}

\usepackage{listings}
\usepackage{censor}
\usepackage[hidelinks]{hyperref} 

\begin{document}
\sloppy

\title{Boosting EfficientNets Ensemble Performance via Pseudo-Labels and Synthetic Images by pix2pixHD for Infection and Ischaemia Classification in Diabetic Foot Ulcers}
\titlerunning{Boosting DFU Classification via Pseudo-Labels and Synthetic Images}
%
\author{Louise Bloch\inst{1,2}$^{,\dagger,*}$\orcidID{0000-0001-7540-4980} \and
Raphael Br\"ungel\inst{1,2}$^{,\dagger}$\orcidID{0000-0002-6046-4048} \and
Christoph M. Friedrich\inst{1,2}\orcidID{0000-0001-7906-0038}}
\authorrunning{L. Bloch et al.}
%

\institute{Department of Computer Science, University of Applied Sciences and Arts Dortmund (FH Dortmund), Emil-Figge-Str. 42, 44227 Dortmund, Germany \and
Institute for Medical Informatics, Biometry and Epidemiology (IMIBE), University Hospital Essen, Hufelandstr. 55, 45122 Essen, Germany
\email{ [louise.bloch,raphael.bruengel,christoph.friedrich]@fh-dortmund.de} \\
$^{\dagger}$ These authors contributed equally to this work\\
$^{*}$ Corresponding author}

\maketitle              
\begin{abstract}
Diabetic foot ulcers are a common manifestation of lesions on the diabetic foot, a syndrome acquired as a long-term complication of diabetes mellitus. Accompanying neuropathy and vascular damage promote acquisition of pressure injuries and tissue death due to ischaemia. Affected areas are prone to infections, hindering the healing progress. The research at hand investigates an approach on classification of infection and ischaemia, conducted as part of the Diabetic Foot Ulcer Challenge (DFUC) 2021. Different models of the EfficientNet family are utilized in ensembles. An extension strategy for the training data is applied, involving pseudo-labeling for unlabeled images, and extensive generation of synthetic images via pix2pixHD to cope with severe class imbalances. The resulting extended training dataset features $8.68$ times the size of the baseline and shows a real to synthetic image ratio of $1:3$. Performances of models and ensembles trained on the baseline and extended training dataset are compared. Synthetic images featured a broad qualitative variety. Results show that models trained on the extended training dataset as well as their ensemble benefit from the large extension. F1-Scores for rare classes receive outstanding boosts, while those for common classes are either not harmed or boosted moderately. A critical discussion concretizes benefits and identifies limitations, suggesting improvements. The work concludes that classification performance of individual models as well as that of ensembles can be boosted utilizing synthetic images. Especially performance for rare classes benefits notably.
\end{abstract}


\section{Introduction} 
\label{sec:introduction}

In 2019 there was an estimated amount of 463 million diabetes mellitus cases ($9.3~\%$ of the world's population) \cite{Saeedi_2019}. This number is expected to rise up to 578 million cases ($10.2~\%$) until 2030 \cite{Saeedi_2019}. Associated with the disease is the diabetic foot syndrome, a long-term complication that can manifest with neuropathy and ischaemia. Without proper monitoring and care, diabetic foot ulcers (DFUs) may arise from these, which have an estimated global prevalence of $6.3~\%$ in diabetics \cite{zhang2016dfu_prevalence}. Impaired wound healing \cite{falanga2005diabeticwoundhealing} and common complications such as infections \cite{siddiqui2010chronic_wound_infection} facilitate chronification, hence regular and attentive screening and documentation are necessitated. Deficient care can prolong treatment, cause aggravation, and ultimately make amputations necessary. Beside a resulting harsh impact on the quality of life, amputation wounds are again prone to complications.

To support overburdened caregivers and facilitate best practices, machine learning-based applications are a key technology. These enable automation of time-consuming tasks and provision of decision support at the point-of-care. This includes the early and certain recognition of adverse shifts in the wound healing progress such as infection and ischaemia. The DFU Challenge (DFUC) is a series of academic challenges that address tasks related to DFU care to enable a broad comparison of detection \cite{dfuc2020}, classification \cite{dfuc2021}, and segmentation \cite{dfuc2022} methods as well as to evaluate the state of the art \cite{yap2021detection} for potential applications.

The work at hand presents a contribution to the DFUC 2021 \cite{dfuc2021} on classification of infection and ischaemia in DFU images. It uses an EfficientNets \cite{EfficientNet} ensemble that achieved the \nth{2} place. Its models were trained on an extended and class-balanced dataset. This was established by via pseudo-labeling of unlabeled images and, as a novelty in DFU classification, via class-individual generation of synthetic images using pix2pixHD \cite{wang2018}.
Related work on DFU classification was conducted by \cite{alzubaidi2019dfu_qutnet,das2021dfu_spnet,goyal2020dfunet} to discriminate healthy and abnormal skin. Recent and strongly related work on classification of infection and ischaemia in DFU was addressed by \cite{das2021recognition,goyal2020dfunet}. Benchmark results for the DFUC 2021 were presented in \cite{10.1109/BHI50953.2021.9508563}. Generation of synthetic wound images was priorly addressed by \cite{sarp2021woundgan,zhang2018woundgan}, yet not specifically for DFU.

The manuscript consent is organized as follows: In Section \ref{sec:data_and_methods} descriptions on used data, methods, and the experiment environment are covered. The approach followed as well as the used experiment setup are elaborated in Section \ref{sec:approach}. Results achieved during the challenge are presented in Section \ref{sec:results} featuring visualizations for explainability, discriminating those without and with the use of pseudo-labels and synthetic images. Section \ref{sec:discussion} provides a critical discussion on the approach, results, and limitations. Eventually, Section \ref{sec:conclusion} summarizes results and draws conclusions on the potential of the presented approach.


\section{Data and Methods}
\label{sec:data_and_methods}

In the following, the DFUC 2021 challenge dataset with its modalities is described. Further,  EfficientNets and pix2pixHD as used methods as well as the environment experiments were performed in are elaborated.


\subsection{Diabetic Foot Ulcer Challenge 2021 Dataset} 

\begin{figure}[ht!]
    \centering
    \begin{subfigure}[b]{0.24\textwidth}
        \centering
        \includegraphics[width=\textwidth]{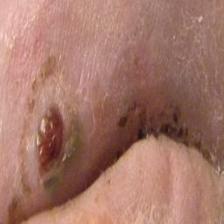}
        \caption{Orig. \texttt{none}}
        \label{fig:DatasetImage_none}
    \end{subfigure}
    \hfill
    \begin{subfigure}[b]{0.24\textwidth}
        \centering
        \includegraphics[width=\textwidth]{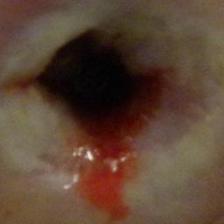}
        \caption{Orig. \texttt{infection}}
        \label{fig:DatasetImage_infection}
    \end{subfigure}
    \hfill
    \begin{subfigure}[b]{0.24\textwidth}
        \centering
        \includegraphics[width=\textwidth]{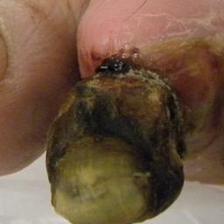}
        \caption{Orig. \texttt{ischaemia}}
        \label{fig:DatasetImage_Ischaemia}
    \end{subfigure}
    \hfill
    \begin{subfigure}[b]{0.24\textwidth}
        \centering
        \includegraphics[width=\textwidth]{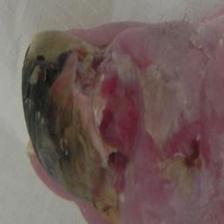}
        \caption{Orig. \texttt{both}}
        \label{fig:DatasetImage_both}
    \end{subfigure}
    \begin{subfigure}[b]{0.24\textwidth}
        \centering
        \includegraphics[width=\textwidth]{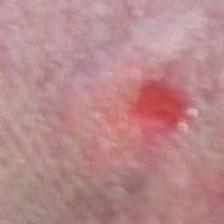}
        \caption{Orig. \texttt{none}}
        \label{fig:DatasetImage_none2}
    \end{subfigure}
    \hfill
    \begin{subfigure}[b]{0.24\textwidth}
        \centering
        \includegraphics[width=\textwidth]{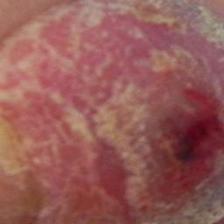}
        \caption{Orig. \texttt{infection}}
        \label{fig:DatasetImage_Infection2}
    \end{subfigure}
    \hfill
    \begin{subfigure}[b]{0.24\textwidth}
        \centering
        \includegraphics[width=\textwidth]{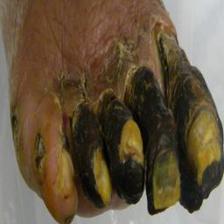}
        \caption{Orig. \texttt{ischaemia}}
        \label{fig:DatasetImage_Ischaemia2}
    \end{subfigure}
    \hfill
    \begin{subfigure}[b]{0.24\textwidth}
        \centering
        \includegraphics[width=\textwidth]{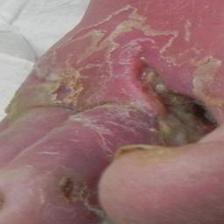}
        \caption{Orig. \texttt{both}}
        \label{fig:DatasetImage_both2}
    \end{subfigure}
    \caption{Examples from the DFUC 2021 dataset for all classes.}
    \label{fig:DFUOrigImages}
\end{figure}

The DFUC 2021 \cite{dfuc2021} dataset \cite{10.1109/BHI50953.2021.9508563} focuses on identification and analysis of infection and ischaemia DFU images. It comprises four classes, showing neither infection nor ischaemia (\texttt{none}), either infection (\texttt{infection}) or ischaemia (\texttt{ischaemia}), or both combined (\texttt{both}). Data was collected from Lancashire Teaching Hospitals\footnote{Lancashire Teaching Hospitals: \url{https://www.lancsteachinghospitals.nhs.uk/}, access 2021-09-22} in a non-laboratory environment. Hence, images comprise flaws such as blurring, poor lighting, and reflection artifacts. Experts extracted patches \cite{10.1109/BHI50953.2021.9508563} with a resolution of $224 \times 224$ px containing DFU regions. The resulting dataset was split into a training and a test part, images of both partitions were augmented to generate additional data, excluding too similar images \cite{10.1109/BHI50953.2021.9508563}. The overall process resulted in $15,683$ images: $5,955$ ($37.97~\%$) labeled training images, $3,994$ ($25.47~\%$) unlabeled training images, and $5,734$ ($36.56~\%$) test images. A validation dataset of $500$ ($8.72~\%$) images was extracted from the test part. The labeled training part comprises $2,555$ ($42.91~\%$) \texttt{infection} images, $227$ ($3.81~\%$) \texttt{ischaemia} images, $621$ ($10.43~\%$) \texttt{both} images, and $2,552$ ($42.85~\%$) \texttt{none} images \cite{10.1109/BHI50953.2021.9508563}. 
Figure \ref{fig:DFUOrigImages} shows examples, provided by the maintainers.

Beside the low resolution of patches and the overlapping class \texttt{both}, the dataset features further obstacles. The risk of information leakage is present due to an unclear generation of original training and test sets which might not be split on the subject level. In addition, the choice of augmented image inclusion can be questioned as whether augmentations should rather be dedicated to challenge contestants, as model selection strategies are impacted by these. 

\subsection{Classification via EfficientNets} 

The EfficientNet\footnote{EfficientNet: \url{https://github.com/mingxingtan/efficientnet}, access 2021-10-03}~\cite{EfficientNet} base model is a classification network developed using a CNN architecture search. The search aims to optimize classification models for performance (measured in accuracy) and training time (measured in Floating Point Operations Per Second (FLOPS)) in parallel.
To increase image resolution, model depth, and model width, this base model is gradually scaled up using a uniform balance. All models of the EfficientNet family (EfficientNet-B0 up to EfficientNet-B7) achieved state-of-the-art performances on the ImageNet \cite{Deng2009} classification task using smaller and faster model architectures~\cite{EfficientNet}. 


\subsection{Image Synthesis via pix2pixHD} 

The pix2pixHD\footnote{pix2pixHD: \url{https://github.com/NVIDIA/pix2pixHD}, access 2021-09-12} \cite{wang2018} framework enables photo-realistic high-resolution image synthesis and image-to-image translation for images up to $2048 \times 1024$ px. It represents a refined version of pix2pix \cite{isola2017}, based on a conditional \cite{mirza2014cgans} Generative Adversarial Network (GAN) \cite{goodfellow2014gans} architecture, combining a novel and more robust adversarial learning objective with a multi-scale generator/discriminator \cite{wang2018}. Hereby, it addresses the problem of lacking details and realistic textures for high resolutions \cite{isola2017,wang2018}. pix2pixHD further features interactive semantic manipulation for objects on an instance level as well as generation of different synthetic images for a single input \cite{wang2018}. Beside the use of semantic label masks, it also allows training and generation via edge masks in a zero-class mode.


\subsection{Experimental Environment} 

Experiments were conducted on NVIDIA\textsuperscript{\textregistered} V100\footnote{V100: \url{https://www.nvidia.com/en-us/data-center/v100/}, access 2021-09-13} tensor core Graphical Processing Units (GPUs) with 16 GB memory. These were part of an NVIDIA\textsuperscript{\textregistered} DGX-1\footnote{DGX-1: \url{https://www.nvidia.com/en-us/data-center/dgx-1/}, access 2021-09-13}, a supercomputer specialized for deep learning. The operating system was Ubuntu Linux\footnote{Ubuntu Linux: \url{https://ubuntu.com/}, access 2021-07-10} in version \texttt{20.04.2 LTS (Focal Fossa)}, the driver version was \texttt{450.119.04}, and the used Compute Unified Device Architecture (CUDA) version was \texttt{10.1}. The execution environment was an NVIDIA\textsuperscript{\textregistered}-optimized\footnote{NVIDIA\textsuperscript{\textregistered}-Docker: \url{https://github.com/NVIDIA/nvidia-docker}, access 2021-07-10} Docker\footnote{Docker: \url{https://www.docker.com/}, access 2021-07-10} \cite{merkel2014docker} container engine, running a Deepo\footnote{Deepo: \url{https://github.com/ufoym/deepo}, access 2021-09-22} image for a quick setup. Unless stated otherwise, experiments were conducted on a single GPU.


\section{Approach}
\label{sec:approach}

\begin{figure}[ht!]
    \centering
    \includegraphics[width=1.0\linewidth]{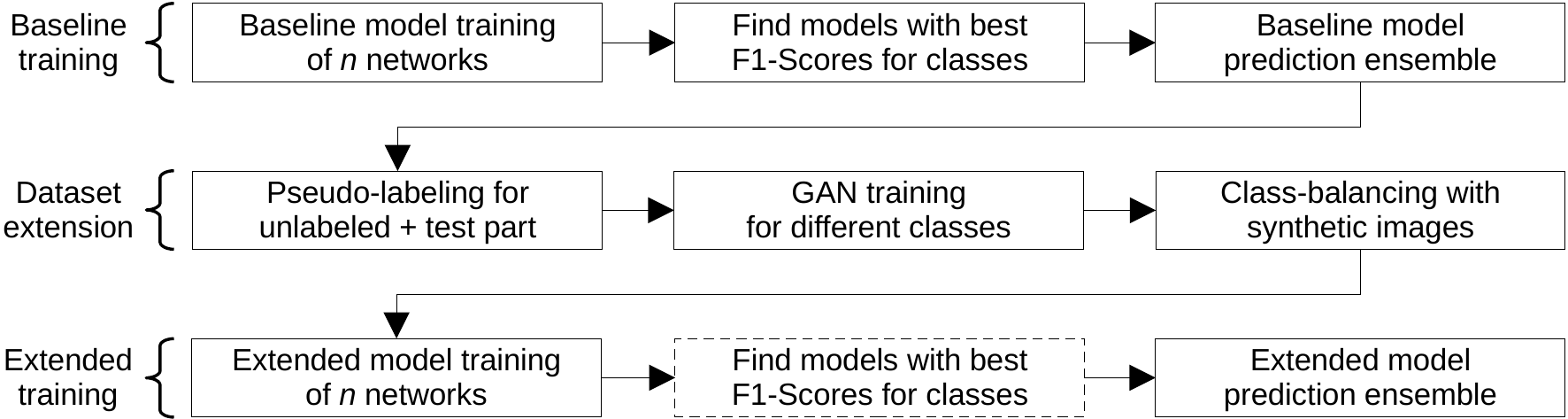}
    \caption{Implemented three-phase workflow: Training with baseline data, extension of baseline data, and training with extended data. In the third phase, no F1-Score evaluation (dashed box) was possible due to expiration of the validation phase.}
    \label{fig:workflow}
\end{figure}

In the following, the implemented approach visualized in Figure \ref{fig:workflow} and divided into three phases is elaborated. In the baseline phase, different deep learning-based models were trained on the baseline training dataset and the best performing models, all of the EfficientNet family, were combined to a prediction ensemble. The average ensemble generated pseudo-labels for the unlabeled and test part of the DFUC 2021 dataset to extend available training data. The baseline training dataset and highly confident pseudo-labels were then used to train class-individual pix2pixHD models, utilized to generate synthetic images for class-balancing. Based on this final extended training dataset, comprising the baseline training dataset, pseudo-labels for unlabeled and test part images, and synthetic images, different models of the EfficientNet family with the initially best performing configuration were trained and merged to a prediction ensemble.

\subsection{Baseline Models and Prediction Ensemble} 

During the validation stage of the challenge, explorative experiments were executed to investigate the performances of different deep learning-based models, including EfficientNets \cite{EfficientNet}, EfficientNet-v2 \cite{EfficientNetv2}, Vision Transformers \cite{VisionTransformers} and ResNet 101 \cite{ResNet}. Those models were loaded using the Python package PyTorch image models (\texttt{timm}) \cite{rw2019timm} and trained using the Python package PyTorch \cite{PyTorch} on the original training dataset. Further experiments were executed using different learning rates, numbers of epochs, optimizers, oversampling strategies, and a step based learning-rate scheduler. All models were pre-trained using the ImageNet-1k or ImageNet-21k dataset. For some models, a warm-up phase was implemented to first train the added classification layers. 
Cross-entropy loss was used for all experiments. For each model, the highest mini-batch size was determined. Mixed precision \cite{Micikevicius2018} was implemented to decrease memory requirements during the training process and consequently increase mini-batch size. The input images were $224 \times 224$ px.  
Two augmentation pipelines -- one baseline pipeline and one extended pipeline -- were implemented using the Albumentations \cite{Buslaev_2020} package. The baseline augmentation pipeline consists of basic augmentations: resizing, random cropping, vertical and horizontal flipping, geometrical shifting, scaling and rotation, an RGB shift, random brightness contrast and image normalization. The extended augmentation pipeline included resizing, random cropping, vertical and horizontal flipping, geometrical shifting, scaling and rotation, random brightness contrast, blurring and median blurring, downscaling, elastic transforms, optical distortions, grid distortions, and image normalization.

For all models, a 5-fold cross-validation (CV) was implemented. However, since the baseline training dataset contains augmented images \cite{10.1109/BHI50953.2021.9508563}, training and validation sets of CV-splits were not independent, leading to overestimated model performances. Those baseline models that reached the best class F1-Scores during the validation stage were combined to an average ensemble. Baseline model parameters are summarized in Table \ref{tab:classificationparameters}, results during the validation stage are summarized in Table \ref{tab:ResultsValidation}. To improve model performances and generalizability \cite{Krizhevsky2012}, averaging was implemented without weights. The average ensemble was used to generate pseudo-labels for unlabeled images of training and test parts.

\begin{table}[ht!]
	\begin{center}
		\caption{Used hyperparameters to train the baseline models. All models used a dropout ratio of 0.3, were pre-trained for the ImageNet-1k dataset, and used an image size of $224 \times 224$ px.}\label{tab:classificationparameters}
			\begin{tabularx}{\textwidth}{|l|X|X|X|X|}
			\hline
			\textbf{Parameter} & \textbf{B$_1$} & \textbf{B$_2$} & \textbf{B$_3$} & \textbf{B$_4$}\\
			\hline
			\hline	
		EfficientNet model architecture & B1 & B0&B2 & B1\\
			Epochs warm-up & 0 & 0 & 0 & 3\\
			Learning rate warm-up & No & No & No & $10^{-2}$ \\
			Epochs training & 30 & 100 & 30 & 47 \\
			Learning rate training & $10^{-4}$ & $10^{-4}$ & $10^{-4}$ & $10^{-4}$ \\
			Batch size & 225 & 300 & 225 & 225\\
			Oversampling & No & Yes & No & No \\
			Augmentations & Baseline&Extended&Baseline&Baseline\\
			\hline
			Optimizer & Adam & Adam & Adam & RMSprop\\
			\hline
			Learning rate scheduler & No & No & Step & Step \\
			Step size & No & No & 10 & 10 \\
			Gamma & No & No & 0.1 & 0.1\\
			\hline
		\end{tabularx}		
	\end{center}
\end{table}

\begin{table}[ht!]
	\begin{center}
		\caption{Official classification results of the baseline models for the validation part of the dataset: Macro, weighted average (WA), and class F1-Scores (F1) as well as the Accuracy. Best results are highlighted.}\label{tab:ResultsValidation}
		\begin{tabularx}{\textwidth}{|X|r|r|r|r|r|r|r|r|}
			\hline
			\textbf{} & \textbf{\texttt{none}} & \textbf{\texttt{infection}} & \textbf{\texttt{ischaemia}} & \textbf{\texttt{both}}&\textbf{}&\textbf{WA}&\textbf{macro} \\
			\textbf{Model} & \textbf{F1 \%} & \textbf{F1 \%} & \textbf{F1 \%} & \textbf{F1 \%} &\textbf{Acc. \%}&\textbf{F1 \%}& \textbf{F1 \%}\\
			\hline\hline
			B$_1$&71.02&58.64&35.90&\textbf{56.18}&63.60&\textbf{63.04}&\textbf{55.43}\\
			B$_2$&68.57&57.29&\textbf{39.13}&52.50&61.60&61.15&54.37\\
			B$_3$&\textbf{72.41}&52.00&38.89&45.65&61.60&59.79&52.24\\
			B$_4$&71.05&\textbf{59.85}&37.04&49.41&\textbf{63.80}&\textbf{63.04}&54.34\\\hline
		\end{tabularx}
	\end{center}
\end{table}

\subsection{Pseudo-Labeling and Synthetic Image Generation} 

In the second phase, the baseline training dataset was extended in two steps: (i) Creation of pseudo-labels for not yet labeled images for initial extension, in particular for the underrepresented classes \texttt{ischaemia} and \texttt{both}, and (ii) generation of synthetic images to extend training data as well as to cope with class imbalances. Details of the resulting class distribution are listed in Table \ref{tab:pseudolabels_synthetic} and further elaborated in the following.

For pseudo-labeling the model ensemble created in workflow phase 1 was used to infer predictions for the unlabeled and test part of the dataset. In sum, both dataset parts comprised $9,728$ images ($3,994$ unlabeled, $5,734$ test). To only use quite confident predictions, a confidence threshold of $70~\%$ for a single class was set as condition to ascribe an image to it. This was done to exclude rather uncertain predictions that would have been more likely to represent false-positive cases, having a negative impact on the classification performance of models trained on the extended dataset. 
A total of $6,961$ predictions fulfilled the set requirement and were considered as pseudo-labeled training data extension. This way, the amount of images of the \texttt{ischaemia} class could be increased by $189$ ($+83.26~\%$), and that of the \texttt{both} class by $321$ ($+51.69~\%$). The amount of images for the \texttt{none} and \texttt{infection} classes could be increased as well by $4,348$ ($+170.38~\%$) and $2,103$ ($+82.31~\%$). Yet, their extension was less crucial for the second extension step due to an already sufficient amount of images. After the first step of pseudo-labeling, the \texttt{none} class comprised $6,900$ images, \texttt{infection} $4,658$ images, \texttt{ischaemia} $416$ images, and \texttt{both} $942$ images.

\begin{figure}[ht!]
    \centering
    \begin{subfigure}[b]{0.24\textwidth}
        \centering
        \includegraphics[width=\textwidth]{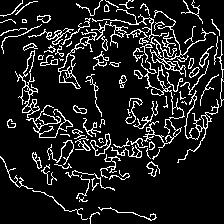}
        \caption{Mask \texttt{none}}
        \label{fig:gan_balance_a}
    \end{subfigure}
    \hfill
    \begin{subfigure}[b]{0.24\textwidth}
        \centering
        \includegraphics[width=\textwidth]{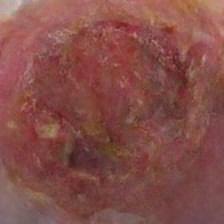}
        \caption{Syn. \texttt{infection}}
        \label{fig:gan_balance_b}
    \end{subfigure}
    \hfill
    \begin{subfigure}[b]{0.24\textwidth}
        \centering
        \includegraphics[width=\textwidth]{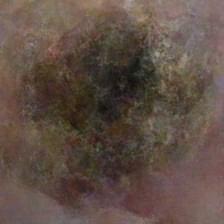}
        \caption{Syn. \texttt{ischaemia}}
        \label{fig:gan_balance_c}
    \end{subfigure}
    \hfill
    \begin{subfigure}[b]{0.24\textwidth}
        \centering
        \includegraphics[width=\textwidth]{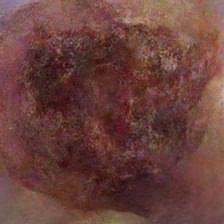}
        \caption{Syn. \texttt{both}}
        \label{fig:gan_balance_d}
    \end{subfigure}
    \caption{Examples for synthetic images generated via a mask of the \texttt{none} class for the \texttt{infection}, \texttt{ischaemia}, and \texttt{both} classes.}
    \label{fig:gan_balance}
\end{figure}

For synthetic image generation, individual pix2pixHD models for each class had to be created. As no area masks with regions of interest were available for images, edge masks were created for images of the extended dataset using the Canny edge detection algorithm \cite{canny1986} implemented in ImageMagick\footnote{ImageMagick: \url{https://github.com/ImageMagick/ImageMagick}, access 2021-09-22} version \texttt{6.9.10-23 Q16 x86\_64 20190101}. The default parameterization was used, setting the radius to $0$, the standard deviation to $1$, and the percent level range to $[10, 30]$. Resulting edge masks enabled training in a zero-class mode, considering the whole image content with the aid of a respective sketch as a support structure. Individual pix2pixHD models were then trained on class-specific splits of the training dataset extended with pseudo-labeled images from the first step. Used parameters and settings are listed in Table \ref{tab:p2phdtrain}. The chosen batch size was the maximum possible amount of images, limited by the GPU RAM, yet increased by using mixed precision. The default learning rate of $2\cdot 10^{-4}$ was raised to $3\cdot 10^{-4}$, as instabilities\footnote{pix2pixHD artifacts: \url{https://github.com/NVIDIA/pix2pixHD/issues/46}, access 2021-09-11}, occurring during early stages of training, were less likely to persist. The amount of epochs with the initial and decaying learning rate was chosen manually by observing intermediate results during training when synthetic images were decided to be sufficiently detailed and convincing.
Trained models were then used to generate synthetic images. To cope with the considerable class imbalance, for each class synthetic images were created using the edge masks of all other classes. I.e., the $6,900$ given \texttt{none} images were extended generating further $6,016$ synthetic images via the \texttt{none} model, using the $4,658$ \texttt{infection}, $416$ \texttt{ischaemia}, and $942$ \texttt{both} class edge masks and vice versa. Figure \ref{fig:gan_balance} illustrates the translation of a single edge mask of the \texttt{none} class (Figure \ref{fig:gan_balance_a}) to three synthetic images of the \texttt{infection}, \texttt{ischaemia}, and \texttt{both} classes (Figure \ref{fig:gan_balance_b}, Figure \ref{fig:gan_balance_c}, and Figure \ref{fig:gan_balance_d}). Hence, after the second step of synthetic image extension each class comprised $12,916$ labeled images, summing up to $51,664$ training images including baseline, pseudo-labeled and synthetic images.

\begin{table}[ht!]
	\begin{center}
	\caption{Proportions of the extended training dataset after two extensions.}
		\label{tab:pseudolabels_synthetic}
			\begin{tabularx}{\textwidth}{|X|r|r|r||l|}
			\hline
			\textbf{}& \textbf{Baseline}&\textbf{Pseudo-label} & \textbf{Syn. image}&\\
			\textbf{Class} & \textbf{training data} & \textbf{extension} & \textbf{extension} & \textbf{$\Sigma$} \\
			\hline
			\hline	
			\texttt{none} & $2,552$ ($4.94 \;\%$) & $4,348$ ($8.42 \;\%$) & $6,016$ ($11.64 \;\%$) & $12,916$ ($25.00 \;\%$) \\
			\texttt{infection} & $2,555$ ($4.95 \;\%$) & $2,103$ ($4.07 \;\%$) & $8,258$ ($15.98 \;\%$) & $12,916$ ($25.00 \;\%$) \\
			\texttt{ischaemia} & 227 ($0.44 \;\%$) & $189$ ($0.37 \;\%$) & $12,500$ ($24.19 \;\%$) & $12,916$ ($25.00 \;\%$) \\
			\texttt{both} & $621$ ($0.12 \;\%$) & $321$ ($0.62 \;\%$) & $11,974$ ($23.18 \;\%$) & $12,916$ ($25.00 \;\%$) \\
			\hline
			\hline
			$\Sigma$ & $5,955$ ($11.53 \;\%$) & $6,961$ ($13.47 \;\%$) & $38,748$ ($75.00 \;\%$) & $51,664$ ($100.00 \;\%$) \\
			\hline
		\end{tabularx}
	\end{center}
\end{table}

\begin{table}[ht!]
	\begin{center}
		\caption{pix2pixHD parameters used for individual class model training.}
		\label{tab:p2phdtrain}
			\begin{tabularx}{\textwidth}{|l|X|X|X|X|}
			\hline
			\textbf{Parameter/Setting} & \texttt{none} & \texttt{infection} & \texttt{ischaemia} & \texttt{both} \\
			\hline
			\hline
			Number of classes & $0$ & $0$ & $0$ & $0$ \\
			Mixed precision & Yes & Yes & Yes & Yes \\
			Batch size & $48$ & $48$ & $48$ & $48$ \\
			Learning rate & $3\cdot 10^{-4}$ & $3\cdot 10^{-4}$ & $3\cdot 10^{-4}$ & $3\cdot 10^{-4}$ \\
			Epochs with initial learning rate & 50 & 50 & 200 & 200 \\
			Epochs with decaying learning rate & 100 & 100 & 400 & 400 \\
			Load/fine size & $224$ px & $224$ px & $224$ px & $224$ px \\
			Resize/crop & No & No & No & No \\
			Instance maps & No & No & No & No \\
			\hline
		\end{tabularx}
	\end{center}
\end{table}

\subsection{Extended Models and Prediction Ensemble} 

Based on the synthetic images generated in the second phase of the workflow, three models were trained with the deep learning-based classification pipeline of phase 1. To increase the mini-batch size and decrease the training time, the pipeline was trained using four GPU cores. Due to the expiration of the challenge's validation phase, no further hyperparameter tuning was performed. Instead, the hyperparameters of Baseline model 1 were used because it reached the best macro F1-Score during the validation stage. The extended models were trained using the same hyperparameters but the EfficientNet-B0, EfficientNet-B1 and EfficientNet-B2 classification architectures. 

Finally, unweighted averaging was implemented to create an ensemble of the three models. The average ensemble was used to generate the final predictions.


\section{Results}
\label{sec:results}

Results achieved for the different workflow stages are described in the subsequent sections. Classification results are summarized for the baseline models, the extended models, and their average ensembles. Additionally, the synthetic images generated for the extended training dataset are presented.


\subsection{Baseline Model and Ensemble Performance} 

The classification results reached during an internal 5-fold CV are summarized in Table \ref{tab:internal_results} and the classification results achieved for the test set are summarized in Table \ref{tab:Results}. The best macro F1-Score for a baseline model during CV was $92.11~\%\pm1.35$ for baseline model 2. This model was an EfficientNet-B0 model trained with oversampling and the extended augmentation pipeline and was thus intended to generate more robust predictions. This model reached the best \texttt{infection} F1-Score of $60.25~\%$ for the test dataset. The best test F1-Score of $72.92~\%$ for the \texttt{none} class comparing the baseline models was reached for baseline model 4. This model was an EfficientNet-B1 model trained with a warm-up phase and the RMSprop \cite{Hinton2012RMSprop} optimizer. This model reached a macro F1-Score of $56.40~\%$ and outperformed the remaining baseline models. Considering baseline models, the best test F1-Score for the \texttt{ischaemia} class was $47.50~\%$ reached for Baseline model 3. In comparison to the remaining baseline models, this model achieved the best F1-Score of $48.58~\%$ for the \texttt{both} class. Baseline model 3 was an EfficientNet-B2 model.

The average baseline ensemble reached a CV macro F1-Score of $90.37~\% \pm 1.23$. This result was slightly worse than the score of baseline model 2. For the test dataset, the average baseline ensemble outperformed all individual models for the F1-Score of the \texttt{none}, \texttt{ischaemia} and \texttt{both} classes, as well as for the macro F1-Score. The macro F1-Score of this model was $59.36~\%$.

\begin{table}[ht!]
	\begin{center}
		\caption{Internal 5-fold CV classification results: Macro and class F1-Scores (F1). All scores are given as $\bar{x} \pm \sigma$. Best results are highlighted.}\label{tab:internal_results}
		\begin{tabularx}{\textwidth}{|X|r|r|r|r|r|}
			\hline
			\textbf{}
			& \textbf{\texttt{none}} & \textbf{\texttt{infection}} & \textbf{\texttt{ischaemia}} & \textbf{\texttt{both}}&\textbf{macro} \\
			\textbf{Model}
			& \textbf{F1 \%} & \textbf{F1 \%} & \textbf{F1 \%} & \textbf{F1 \%} & \textbf{F1 \%}\\
			\hline\hline
			B$_1$ & $86.31 \pm 0.21$&$84.70 \pm 0.96$&$82.18 \pm 2.58$&$88.36 \pm 2.56$&$85.39 \pm 1.39$\\
			B$_2$ &\textbf{90.61 $\pm$ 1.36}&\textbf{90.12 $\pm$ 1.30}& \textbf{91.92 $\pm$ 3.23}&\textbf{95.78 $\pm$ 1.47}&\textbf{92.11 $\pm$ 1.35}\\
			B$_3$ &$80.38 \pm 1.27$&$76.24 \pm 1.46$&$67.83 \pm 7.06$&$79.54 \pm 1.56$&$76.00 \pm 2.02$\\
			B$_4$ &$85.96 \pm 0.48$&$84.15 \pm 1.06$&$83.10 \pm 2.86$&$89.57 \pm 0.92$&$85.69 \pm 0.81$\\\hline
			B$_\text{ensemble}$ & $89.50 \pm 0.39$&$88.41 \pm 0.70$&$89.89 \pm 4.66$&$93.70 \pm 0.78$&$90.37 \pm 1.23$\\\hline\hline
			E$_1$ & $85.64 \pm 1.10$&$83.27 \pm 1.45$&$82.84 \pm 3.52$&$87.23 \pm 2.58$&$84.75 \pm 1.18$\\
			E$_2$ & $86.84 \pm 0.59$&$84.28 \pm 0.55$&$84.49 \pm 2.82$&$89.01 \pm 0.54$&$86.15 \pm 0.61$\\
			E$_3$ &$88.20 \pm 0.57$&$85.89 \pm 0.85$&$86.42 \pm 3.87$&$90.42 \pm 1.45$&$87.73 \pm 1.51$\\\hline
			E$_\text{ensemble}$ & $89.15 \pm 0.63$&$87.28 \pm 0.73$&$90.12 \pm 3.82$&$92.70 \pm 1.03$&$89.81 \pm 0.97$\\
			\hline
		\end{tabularx}		
	\end{center}
\end{table}

\begin{table}[ht!]
	\begin{center}
		\caption{Official classification results for the test part of the dataset: Macro, weighted average (WA), and class F1-Scores (F1) as well as the Accuracy. Best results are highlighted.}\label{tab:Results}
		\begin{tabularx}{\textwidth}{|X|r|r|r|r|r|r|r|}
			\hline
			\textbf{}& \textbf{\texttt{none}} & \textbf{\texttt{infection}} & \textbf{\texttt{ischaemia}} & \textbf{\texttt{both}}&\textbf{}&\textbf{WA}&\textbf{macro} \\
			\textbf{Model}& \textbf{F1 \%} & \textbf{F1 \%} & \textbf{F1 \%} & \textbf{F1 \%} & \textbf{Acc. \%} &\textbf{F1 \%} & \textbf{F1 \%}\\
			\hline\hline
			B$_1$&71.14&57.38&41.50&44.54&62.36&61.73&53.64\\
			B$_2$&72.39&\textbf{60.25}&42.49&46.06&64.11&63.70&55.30\\
			B$_3$ &72.86&55.08&47.50&48.58&63.38&62.05&56.00\\
			B$_4$ &	72.92&59.72&46.81&46.15&64.70&63.88&56.40\\\hline
			B$_\text{ensemble}$ & 74.24&59.54&51.67&51.97&66.08&65.06&59.36\\\hline\hline
			E$_1$ & 74.36&58.46&54.02&51.22&65.99&64.66&59.51\\
			E$_2$ & 74.09&59.15&55.49&50.56&66.01&64.85&59.82\\
			E$_3$ & 74.41&59.05&54.30&53.32&66.34&65.13&60.27\\\hline
			E$_\text{ensemble}$ (2nd) &\textbf{74.53}&59.17&\textbf{55.80}&\textbf{53.59}&\textbf{66.57}&\textbf{65.32}&\textbf{60.77}\\
			\hline
		\end{tabularx}		
	\end{center}
\end{table}


\subsection{Synthetic Images for Training Dataset Extension} 

\begin{figure}[ht!]
    \centering
    \begin{subfigure}[b]{0.24\textwidth}
        \centering
        \includegraphics[width=\textwidth]{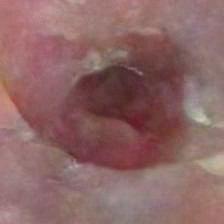}
        \caption{Good \texttt{none}}
        \label{fig:gan_examples_a}
    \end{subfigure}
    \hfill
    \begin{subfigure}[b]{0.24\textwidth}
        \centering
        \includegraphics[width=\textwidth]{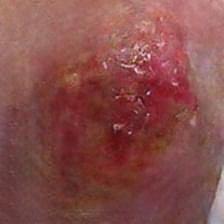}
        \caption{Good \texttt{infection}}
        \label{fig:gan_examples_b}
    \end{subfigure}
    \hfill
    \begin{subfigure}[b]{0.24\textwidth}
        \centering
        \includegraphics[width=\textwidth]{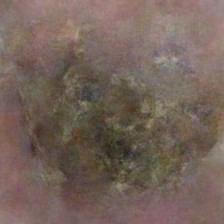}
        \caption{Good \texttt{ischaemia}}
        \label{fig:gan_examples_c}
    \end{subfigure}
    \hfill
    \begin{subfigure}[b]{0.24\textwidth}
        \centering
        \includegraphics[width=\textwidth]{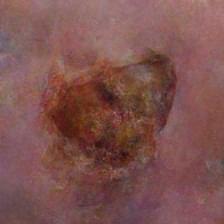}
        \caption{Good \texttt{both}}
        \label{fig:gan_examples_d}
    \end{subfigure}
    \begin{subfigure}[b]{0.24\textwidth}
        \centering
        \includegraphics[width=\textwidth]{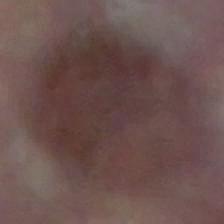}
        \caption{Poor \texttt{none}}
        \label{fig:gan_examples_e}
    \end{subfigure}
    \hfill
    \begin{subfigure}[b]{0.24\textwidth}
        \centering
        \includegraphics[width=\textwidth]{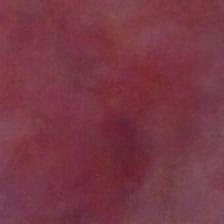}
        \caption{Poor \texttt{infection}}
        \label{fig:gan_examples_f}
    \end{subfigure}
    \hfill
    \begin{subfigure}[b]{0.24\textwidth}
        \centering
        \includegraphics[width=\textwidth]{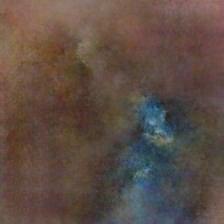}
        \caption{Poor \texttt{ischaemia}}
        \label{fig:gan_examples_g}
    \end{subfigure}
    \hfill
    \begin{subfigure}[b]{0.24\textwidth}
        \centering
        \includegraphics[width=\textwidth]{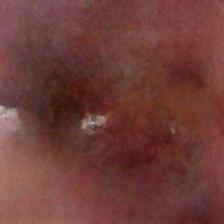}
        \caption{Poor \texttt{both}}
        \label{fig:gan_examples_h}
    \end{subfigure}
    \caption{Examples for generated synthetic images: (a) -- (d) show qualitatively good, (e) -- (h) qualitatively poor results.}
    \label{fig:gan_examples}
\end{figure}

Generated synthetic images showed a broad variety regarding their quality and visual coherence, examples are shown in Figure \ref{fig:gan_examples}. While no realistically looking extremity-like structures such as toes were generated, contents usually resembled less or more convincing isolated ulcerated/ischaemic areas.

Qualitatively good and convincing results incorporated photo-realistic fine details, e.g., depth through multiple layers of skin with scale-like structures (Figure \ref{fig:gan_examples_a}), granulation-like textures with wetness and reflection artifacts (Figure \ref{fig:gan_examples_b}), infection-like localized redness (Figure \ref{fig:gan_examples_b} and Figure \ref{fig:gan_examples_d}), and localized cyanotic respectively necrotic coloring and textures (Figure \ref{fig:gan_examples_c} and Figure \ref{fig:gan_examples_d}). Qualitatively poor results suffered from either unsharp (Figure \ref{fig:gan_examples_e} and Figure \ref{fig:gan_examples_f}) or unconvincing (Figure \ref{fig:gan_examples_g} and Figure \ref{fig:gan_examples_h}) representations. The \texttt{ischaemia} and \texttt{both} models, in particular, trained with few images, were prone to generate less convincing synthetic images, compared to that generated by the \texttt{none} and \texttt{infection} models.

Generated color schemes were usually consistent, yet the \texttt{ischaemia} model tended to include blue areas (Figure \ref{fig:gan_examples_g}), learned from occasional blue backgrounds in the few baseline images of the respective class.

\subsection{Extended Model and Ensemble Performances}

Using the synthetic images, models were trained to improve the results of the average baseline ensemble. Due to time limitations during the challenge, the classification pipelines for these models were not as diverse as the baseline models. However, the three models differed in using multiple scales of the EfficientNet family. Table \ref{tab:internal_results} summarizes the internal results during 5-fold CV and Table \ref{tab:Results} summarizes the official results for the test set. The best macro F1-Score of $87.73~\% \pm1.51$ during CV was reached by the model E$_3$, which used the EfficientNet-B2 architecture. This model reached the best test F1-Score for the \texttt{none} and \texttt{both} classes. As well as a test macro F1-Score of $60.27~\%$ that outperformed the remaining extended models. The best F1-Scores for the \texttt{infection} and \texttt{ischaemia} classes were achieved for model E$_2$, which was an EfficientNet-B1 model. The F1-Score was $59.15~\%$ for the \texttt{infection} class and $55.49~\%$ for the \texttt{ischaemia} class. All extended models outperformed the average baseline ensemble for the macro F1-Score. Increased F1-Scores can be especially noted for the \texttt{ischaemia} class.

The average ensemble outperformed the individual models for the macro F1-Score during CV, the test F1-Score for all classes, as well as for the test macro F1-Score. The macro F1-Score for the average extended ensemble was $60.77~\%$. 

The results of the three best placements are summarized in Table \ref{tab:ResultsChallenge}, the described ensemble model achieved \nth{2} place. This model outperformed the remaining models for the F1-score of the \texttt{ischaemia} class. More precise documentation about the challenge results are summarized in \cite{cassidy2021diabetic}.

\begin{table}[ht!]
	\begin{center}
		\caption{Official classification results for the test part of the dataset and the three best challenge participants: Macro, weighted average (WA), and class F1-Scores (F1) as well as the Accuracy. Best results are highlighted.}\label{tab:ResultsChallenge}
		\begin{tabularx}{\textwidth}{|X|r|r|r|r|r|r|r|}
			\hline
			\textbf{} 
			& \textbf{\texttt{none}} & \textbf{\texttt{infection}} & \textbf{\texttt{ischaemia}} & \textbf{\texttt{both}}&\textbf{}&\textbf{WA}&\textbf{macro} \\
			\textbf{Challenge placement}  
			& \textbf{F1 \%} & \textbf{F1 \%} & \textbf{F1 \%} & \textbf{F1 \%} & \textbf{Acc. \%} &\textbf{F1 \%} & \textbf{F1 \%}\\\hline\hline
			\nth{1} place \cite{galdran2021convolutional}&\textbf{75.74}&63.88&52.82&\textbf{56.19}&\textbf{68.56}&\textbf{68.01}&\textbf{62.16}\\
			\nth{2} place (this work) 
			&74.53&59.17&\textbf{55.80}&53.59&66.57&65.32&60.77\\
			\nth{3} place &71.57&\textbf{67.14}&45.74&53.90&67.11&67.14&59.59\\
			\hline
		\end{tabularx}		
	\end{center}
\end{table}


\subsection{Local Interpretable Model-agnostic Explanations (LIME)}

\begin{figure}[ht!]
    \centering
    \begin{subfigure}[b]{0.10\textwidth}
        \centering
        \includegraphics[width=\textwidth]{figures/DFUCImagesForPaper/none1.jpg}\\\ \\\ 
        \includegraphics[width=\textwidth]{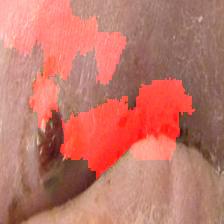}
        \includegraphics[width=\textwidth]{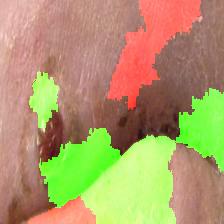}
        \includegraphics[width=\textwidth]{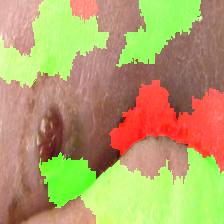} 
        \includegraphics[width=\textwidth]{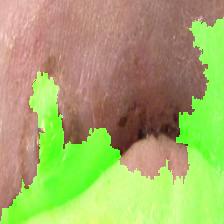}\\\ 
        \includegraphics[width=\textwidth]{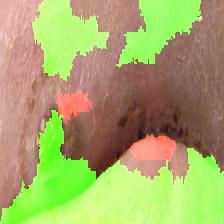}\\\ \\\ 
        \includegraphics[width=\textwidth]{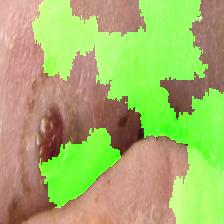}
        \includegraphics[width=\textwidth]{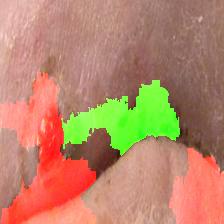}
        \includegraphics[width=\textwidth]{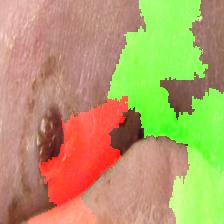}\\\ 
        \includegraphics[width=\textwidth]{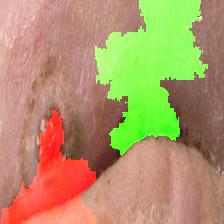}
        \caption{I$_1$}
        \label{fig:lime_i1}
    \end{subfigure}
    \hfill
    \begin{subfigure}[b]{0.10\textwidth}
        \centering
        \includegraphics[width=\textwidth]{figures/DFUCImagesForPaper/none2.jpg}\\\ \\\ 
        \includegraphics[width=\textwidth]{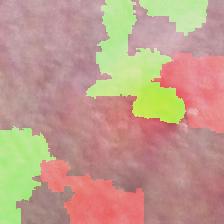} 
        \includegraphics[width=\textwidth]{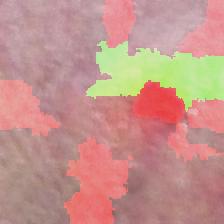} 
        \includegraphics[width=\textwidth]{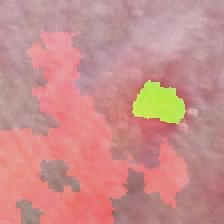} 
        \includegraphics[width=\textwidth]{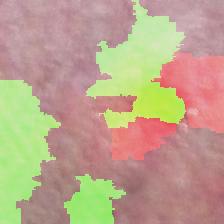}\\\ 
        \includegraphics[width=\textwidth]{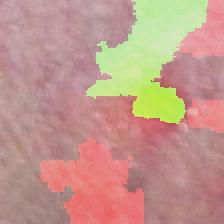} \\\ \\\ 
        \includegraphics[width=\textwidth]{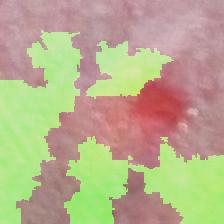} 
        \includegraphics[width=\textwidth]{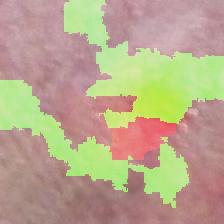} 
        \includegraphics[width=\textwidth]{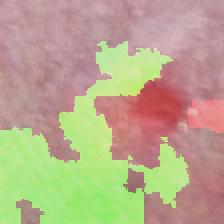}\\\  
        \includegraphics[width=\textwidth]{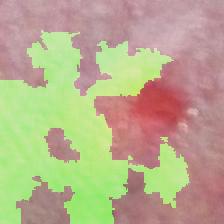}
        \caption{I$_2$}
        \label{fig:lime_i2}
    \end{subfigure}
    \hfill
    \begin{subfigure}[b]{0.10\textwidth}
        \centering
        \includegraphics[width=\textwidth]{figures/DFUCImagesForPaper/infection1.jpg}\\\ \\\ 
        \includegraphics[width=\textwidth]{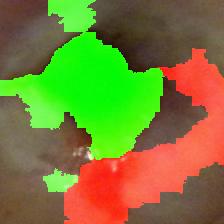} 
        \includegraphics[width=\textwidth]{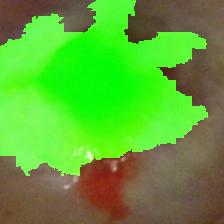} 
        \includegraphics[width=\textwidth]{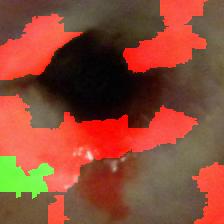} 
        \includegraphics[width=\textwidth]{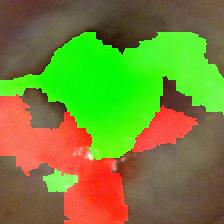}\\\ 
        \includegraphics[width=\textwidth]{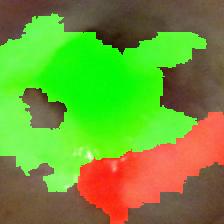}\\\ \\\ 
        \includegraphics[width=\textwidth]{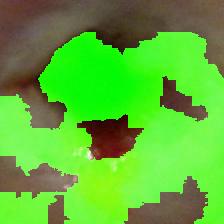} 
        \includegraphics[width=\textwidth]{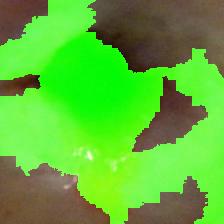}
        \includegraphics[width=\textwidth]{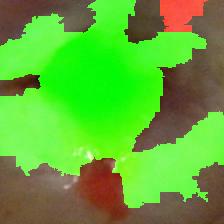}\\\ 
        \includegraphics[width=\textwidth]{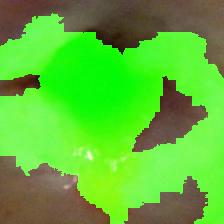}
        \caption{I$_3$}
        \label{fig:lime_i3}
    \end{subfigure}
    \hfill
    \begin{subfigure}[b]{0.10\textwidth}
        \centering
        \includegraphics[width=\textwidth]{figures/DFUCImagesForPaper/infection2.jpg}\\\ \\\ 
        \includegraphics[width=\textwidth]{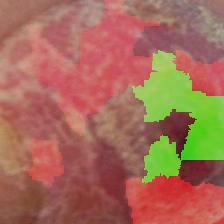} 
        \includegraphics[width=\textwidth]{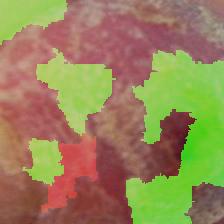} 
        \includegraphics[width=\textwidth]{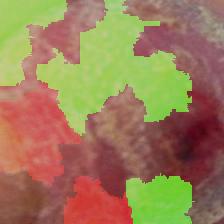} 
        \includegraphics[width=\textwidth]{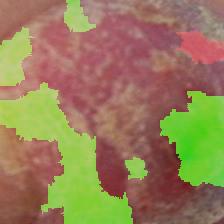}\\\ 
        \includegraphics[width=\textwidth]{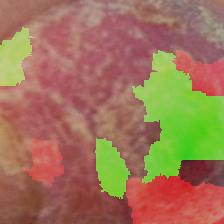}\\\ \\\ 
        \includegraphics[width=\textwidth]{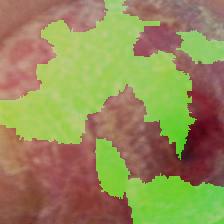}
        \includegraphics[width=\textwidth]{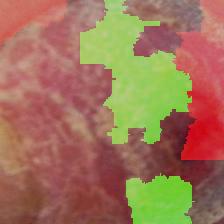}
        \includegraphics[width=\textwidth]{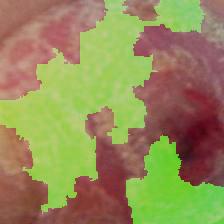}\\\ 
        \includegraphics[width=\textwidth]{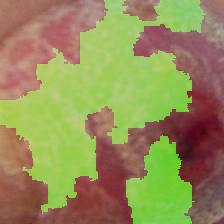}
        \caption{I$_4$}
        \label{fig:lime_i4}
    \end{subfigure}
    \hfill
    \begin{subfigure}[b]{0.10\textwidth}
        \centering
        \includegraphics[width=\textwidth]{figures/DFUCImagesForPaper/ischaemia1.jpg}\\\ \\\ 
        \includegraphics[width=\textwidth]{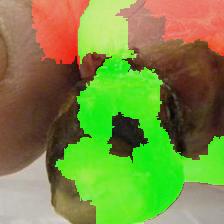} 
        \includegraphics[width=\textwidth]{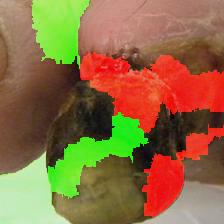} 
        \includegraphics[width=\textwidth]{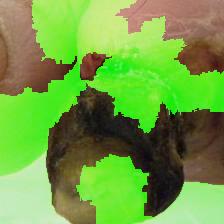} 
        \includegraphics[width=\textwidth]{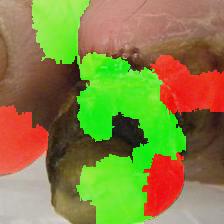}\\\ 
        \includegraphics[width=\textwidth]{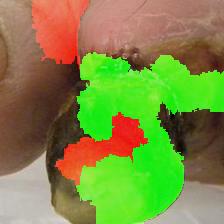}\\\ \\\ 
        \includegraphics[width=\textwidth]{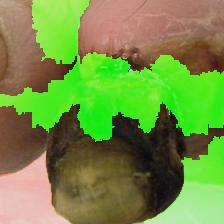} 
        \includegraphics[width=\textwidth]{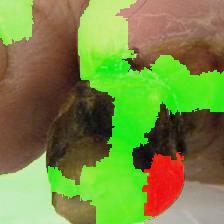} 
        \includegraphics[width=\textwidth]{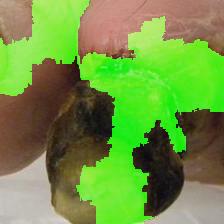}\\\ 
        \includegraphics[width=\textwidth]{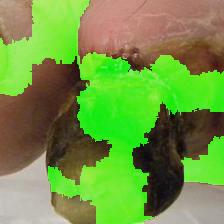}
        \caption{I$_5$}
        \label{fig:lime_i5}
    \end{subfigure}
    \hfill
    \begin{subfigure}[b]{0.10\textwidth}
        \centering
        \includegraphics[width=\textwidth]{figures/DFUCImagesForPaper/ischaemia2.jpg}\\\ \\\ 
        \includegraphics[width=\textwidth]{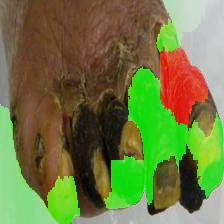}
        \includegraphics[width=\textwidth]{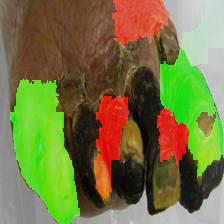} 
        \includegraphics[width=\textwidth]{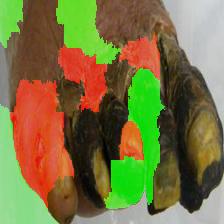} 
        \includegraphics[width=\textwidth]{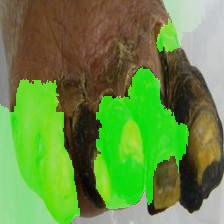}\\\ 
        \includegraphics[width=\textwidth]{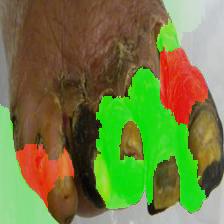}\\\ \\\ 
        \includegraphics[width=\textwidth]{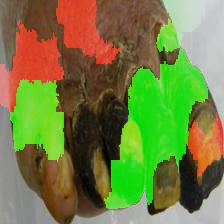} 
        \includegraphics[width=\textwidth]{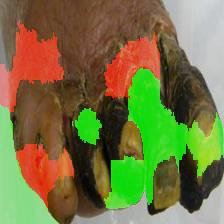} 
        \includegraphics[width=\textwidth]{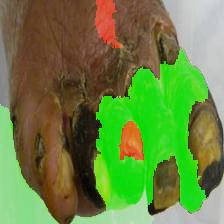}\\\ 
        \includegraphics[width=\textwidth]{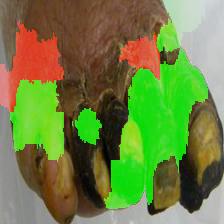}
        \caption{I$_6$}
        \label{fig:lime_i6}
    \end{subfigure}
    \hfill
    \begin{subfigure}[b]{0.10\textwidth}
        \centering
        \includegraphics[width=\textwidth]{figures/DFUCImagesForPaper/both1.jpg}\\\ \\\ 
        \includegraphics[width=\textwidth]{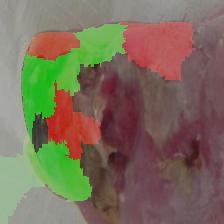} 
        \includegraphics[width=\textwidth]{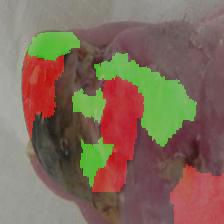} 
        \includegraphics[width=\textwidth]{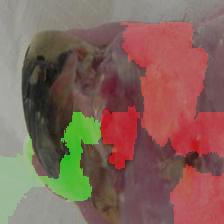} 
        \includegraphics[width=\textwidth]{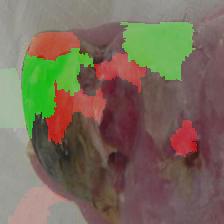}\\\ 
        \includegraphics[width=\textwidth]{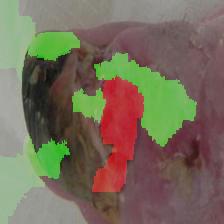}\\\ \\\ 
        \includegraphics[width=\textwidth]{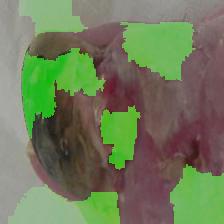} 
        \includegraphics[width=\textwidth]{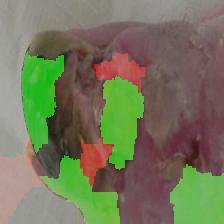} 
        \includegraphics[width=\textwidth]{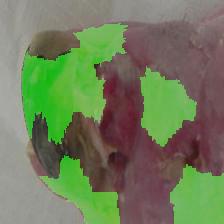}\\\ 
        \includegraphics[width=\textwidth]{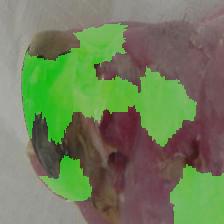}
        \caption{I$_7$}
        \label{fig:lime_i7}
    \end{subfigure}
    \hfill
    \begin{subfigure}[b]{0.10\textwidth}
        \centering
        \includegraphics[width=\textwidth]{figures/DFUCImagesForPaper/both2.jpg}\\\ \\\ 
        \includegraphics[width=\textwidth]{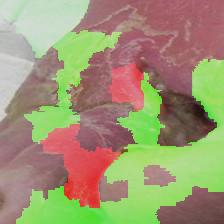} 
        \includegraphics[width=\textwidth]{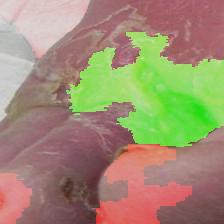} 
        \includegraphics[width=\textwidth]{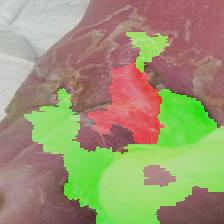} 
        \includegraphics[width=\textwidth]{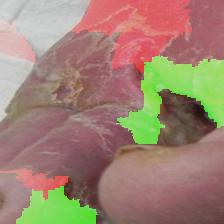}\\\ 
        \includegraphics[width=\textwidth]{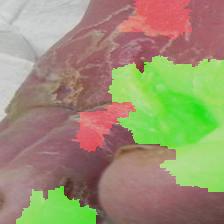}\\\ \\\ 
        \includegraphics[width=\textwidth]{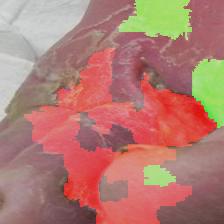} 
        \includegraphics[width=\textwidth]{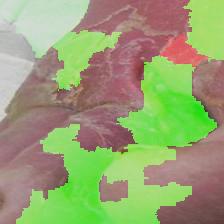} 
        \includegraphics[width=\textwidth]{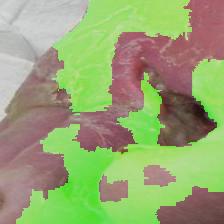}\\\ 
        \includegraphics[width=\textwidth]{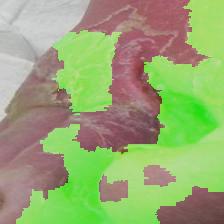}
        \caption{I$_8$}
        \label{fig:lime_i8}
    \end{subfigure}
    \caption{Explainability: (a)--(h) show example images for the classes \texttt{none} (I$_1$, I$_2$), \texttt{infection} (I$_3$, I$_4$), \texttt{ischaemia} (I$_5$, I$_6$), and \texttt{both} (I$_7$, I$_8$) in the first row. The following five rows show LIME decision maps for the four baseline models B$_1$, B$_2$, B$_3$, B$_4$, and their ensemble B$_\text{ensemble}$. The last four rows show respective activation maps for the three extended models E$_1$, E$_2$, E$_3$, and their ensemble E$_\text{ensemble}$. Corresponding class predictions with confidences are listed in Table \ref{tab:lime_predictions}.}
    \label{fig:lime}
\end{figure}

Local Interpretable Model-agnostic Explanations (LIME)\footnote{LIME: \url{https://github.com/marcotcr/lime}, access 2021-11-12} \cite{Ribeiro2016lime} version \texttt{0.2.0.1} are used to visualize the model explanations of example images from the DFUC dataset provided by the maintainers. Per image 3,000 samples were generated to identify the most important superpixels. The 10 most important superpixels for each image are visualized in Figure \ref{fig:lime}, predictions are summarized in Table \ref{tab:lime_predictions}.

\begin{table}[ht!]
	\begin{center}
		\caption{Summary of class predictions of models B$_1$-B$_4$, B$_\text{ensemble}$, E$_1$-E$_3$, and E$_\text{ensemble}$ for example images I$_1$--I$_8$, shown with LIME decision maps in Figure \ref{fig:lime}. False-positive class predictions are highlighted in red.}\label{tab:lime_predictions}
		\begin{tabularx}{\textwidth}{|X|r|r|r|r|r|r|r|r|}
			\hline
			 & \textbf{I$_1$ class} & \textbf{I$_2$ class} & \textbf{I$_3$ class} & \textbf{I$_4$ class} & \textbf{I$_5$ class} & \textbf{I$_6$ class} & \textbf{I$_7$ class} & \textbf{I$_8$ class} \\
			 & \textbf{conf. \%} & \textbf{conf. \%} & \textbf{conf. \%} & \textbf{conf. \%} & \textbf{conf. \%} & \textbf{conf. \%} & \textbf{conf. \%} & \textbf{conf. \%} \\
			\textbf{Model} & (\texttt{none}) & (\texttt{none}) & (\texttt{inf.}) & (\texttt{inf.}) & (\texttt{isc.}) & (\texttt{isc.}) & (\texttt{both}) & (\texttt{both}) \\
			\hline\hline
			B$_1$ & \cellcolor{red!25}\texttt{inf.} & \texttt{none} & \texttt{inf.} & \cellcolor{red!25}\texttt{none} & \cellcolor{red!25}\texttt{both} & \texttt{isc.} & \cellcolor{red!25}\texttt{isc.} & \cellcolor{red!25}\texttt{inf.} \\
			 & \cellcolor{red!25}$61.02$ & $99.99$ & $99.01$ & \cellcolor{red!25}$99.14$ & \cellcolor{red!25}$98.12$ & $63.06$ & \cellcolor{red!25}$83.53$ & \cellcolor{red!25}$99.75$ \\
            B$_2$ & \texttt{none}  & \texttt{none} & \texttt{inf.} & \texttt{inf.} & \texttt{isc.} & \cellcolor{red!25}\texttt{both} & \cellcolor{red!25}\texttt{isc.} & \cellcolor{red!25}\texttt{inf.} \\
			 & $89.55$ & $99.74$ & $99.99$ & $99.98$ & $99.98$ & \cellcolor{red!25}$88.32$ & \cellcolor{red!25}$99.99$ & \cellcolor{red!25}$99.88$ \\ 
			B$_3$ & \cellcolor{red!25}\texttt{inf.} & \texttt{none} & \cellcolor{red!25}\texttt{none} & \cellcolor{red!25}\texttt{none} & \texttt{isc.} & \texttt{isc.} & \cellcolor{red!25}\texttt{inf.} & \cellcolor{red!25}\texttt{inf.} \\
			 & \cellcolor{red!25}$81.64$ & $99.83$ & \cellcolor{red!25}$78.51$ & \cellcolor{red!25}$92.37$ & $61.49$ & $92.83$ & \cellcolor{red!25}$88.43$ & \cellcolor{red!25}$98.57$ \\
			B$_4$ & \cellcolor{red!25}\texttt{inf.} & \texttt{none} & \texttt{inf.} & \cellcolor{red!25}\texttt{none} & \cellcolor{red!25}\texttt{both} & \texttt{isc.} & \cellcolor{red!25}\texttt{inf.} & \cellcolor{red!25}\texttt{inf.} \\
			 & \cellcolor{red!25}$90.30$ & $84.10$ & $97.43$ & \cellcolor{red!25}$96.58$ & \cellcolor{red!25}$87.66$ & $85.47$ & \cellcolor{red!25}$80.20$ & \cellcolor{red!25}$72.77$ \\ 
			\hline
			B$_\text{ensemble}$ & \cellcolor{red!25}\texttt{inf.} & \texttt{none} & \texttt{inf.} & \cellcolor{red!25}\texttt{none} & \cellcolor{red!25}\texttt{both} & \texttt{isc.} & \cellcolor{red!25}\texttt{isc.} & \cellcolor{red!25}\texttt{inf.}\\
			 & \cellcolor{red!25}$60.85$ & $95.92$ & $78.81$ & \cellcolor{red!25}$72.03$ & \cellcolor{red!25}$55.01$ & $63.26$ & \cellcolor{red!25}$47.05$ & \cellcolor{red!25}$92.74$ \\ 
			 \hline\hline
			E$_1$& \texttt{none} & \texttt{none} & \texttt{inf.} & \cellcolor{red!25}\texttt{none} & \cellcolor{red!25}\texttt{both} & \texttt{isc.} & \texttt{both} & \cellcolor{red!25}\texttt{inf.} \\
			 & $91.95$ & $100.00$ & $100.00$ & \cellcolor{red!25}$99.99$ & \cellcolor{red!25}$91.30$ & $99.92$ & $97.27$ & \cellcolor{red!25}$100.00$ \\ 
			E$_2$ & \texttt{none} & \texttt{none} & \texttt{inf.} & \cellcolor{red!25}\texttt{none} & \cellcolor{red!25}\texttt{both} & \texttt{isc.} & \cellcolor{red!25}\texttt{inf.} & \cellcolor{red!25}\texttt{inf.} \\
			 & $99.87$ & $100.00$ & $100.00$ & \cellcolor{red!25}$100.00$ & \cellcolor{red!25}$99.97$ & $88.33$ & \cellcolor{red!25}$94.54$ & \cellcolor{red!25}$100.00$ \\ 
			E$_3$ & \texttt{none} & \texttt{none} & \texttt{inf.} & \cellcolor{red!25}\texttt{none} & \cellcolor{red!25}\texttt{both} & \texttt{isc.} & \texttt{both} & \cellcolor{red!25}\texttt{inf.} \\
			 & $98.66$ & $100.00$ & $100.00$ & \cellcolor{red!25}$100.00$ & \cellcolor{red!25}$99.97$ & $100.00$ & $99.64$ & \cellcolor{red!25}$99.99$ \\ 
			\hline
			E$_\text{ensemble}$ & \texttt{none} & \texttt{none} & \texttt{inf.} & \cellcolor{red!25}\texttt{none} & \cellcolor{red!25}\texttt{both} & \texttt{isc.} & \texttt{both} & \cellcolor{red!25}\texttt{inf.} \\
			 & $96.83$ & $100.00$ & $100.00$ & \cellcolor{red!25}$100.00$ & \cellcolor{red!25}$97.08$ & $96.08$ & $67.46$ & \cellcolor{red!25}$100.00$ \\
			\hline
		\end{tabularx}		
	\end{center}
\end{table}

Superpixels highlighted in green increase the probability of the predicted class (one vs. rest), whereas superpixels highlighted in red decrease the model probability of the predicted class. 

Baseline and extended models as well as their ensembles do not tend to strongly focus on clinically non-relevant areas, such as backgrounds visible in example images I$_5$--I$_8$ for the classes \texttt{ischaemia} and \texttt{both}. Extended models as well as their ensemble also tend to be more certain regarding their predictions, involving greater probability-increasing superpixel areas as can be seen in true-positive predictions for I$_2$, I$_3$, and I$_7$. Yet, this also accounts for false-positive predictions for I$_4$, I$_5$, and I$_8$.


\section{Discussion} 
\label{sec:discussion}

In the following, experiments and results including the model and ensemble development as well as pseudo-labeling and synthetic image generation are discussed. Limitations of the work and in the experiment design are addressed. 


\subsection{Models and Ensembles} 

In this research, state of the art deep learning-based models were used for DFU infection and ischaemia classification. During the validation stage, explorative investigations with different deep-learning architectures and hyperparameters were performed. The best validation F1-Scores for all individual classes were achieved for EfficientNets. More complex models like EfficientNet-v2 and Vision Transformers achieved worse results than the EfficientNets during the validation stage of the challenge. Those best-performing models were combined to an ensemble that outperformed the individual models for the test F1-Score of the \texttt{none}, \texttt{ischaemia} and \texttt{both} classes and for the macro F1-Score. 

The models trained on the extended training dataset outperformed those trained on the baseline training dataset. Extended models achieved outstanding results for the \texttt{ischaemia} class. The averaged extended model ensemble reached the overall best macro F1-Score of $60.77~\%$ and the best class F1-Scores for the \texttt{none}, \texttt{ischaemia}, and \texttt{both} classes. Training on the extended training dataset, therefore, led to considerable improvements for the rare classes \texttt{ischaemia} and \texttt{both}, without harming classification performance for the common classes \texttt{none} and \texttt{infection}. The best model of the challenge benchmark experiments \cite{10.1109/BHI50953.2021.9508563} reached an F1-Score of $55~\%$. The research at hand outperformed this result by $10.49~\%$ ($5.77$ percentage points). The explanations generated via LIME do not indicate that models and ensembles strongly focus on medically irrelevant areas such as backgrounds.


\subsection{Pseudo-Labeling and Synthetic Image Generation} 

Pseudo-labeling of unlabeled data, either for regular dataset extension or self-training approaches, is usually a practicable way to extend available training data, fostering generalization of models. This technique already proved beneficial for a detection task on DFUs \cite{yap2021detection}. In the presented work, creation of pseudo-labeled images allowed to notably increasing the amount of available training data, especially for the rare classes \texttt{ischaemia} and \texttt{both}. Beside images of the test part of the dataset, unlabeled images of the training part were a viable source. The chosen high confidence threshold for inclusion of pseudo-labels is assumed to withheld ingress of the majority of misclassifications, however, no further investigation on this matter was conducted.

The achieved increase of available training images was crucial for class-individual pix2pixHD model training, in particular for the classes \texttt{ischaemia} and \texttt{both}. During initial experiments on the original training part, randomly chosen results of models for these classes solely displayed unconvincing results of poor quality and lacking detail. After the pseudo-label extension, randomly chosen results of re-trained models showed a notably increased quality and higher level of details. Results of the \texttt{none} and \texttt{infection} class models benefited as well, yet initial results usually displayed sufficient detail. The broad variety of results generated by final pix2pixHD models traces back to that of the DFUC 2021 dataset. Missing representations of realistically looking extremities attribute to the overall majority of training images that show small DFUs, solely surrounded by skin. Qualitatively poor and detail-lacking results of \texttt{none} and \texttt{infection} class models were convincing to this extent, that they may be associated with images resulting from poor imaging. However, those of the \texttt{ischaemia} and \texttt{both} class models partially featured unnatural coloring and repeating patterns. This indicates, that even though generation of adequate results is achievable with a few hundred training images, at least a few thousand are required to achieve consistently convincing results with pix2pixHD.

The aggressive approach of class imbalance compensation with massive amounts of synthetic images essentially improved the extended EfficientNets model ensemble performance for the rare classes \texttt{ischaemia} and \texttt{both}. In contrast, performance for the common classes \texttt{none} and \texttt{infection} did not suffer despite considerable amounts of qualitatively poor and potentially unconvincing samples were part of the overall extension. As color schemes of synthetic images were consistent regardless of their quality, beneficial effects may be rather attributable to these than to fine details of synthesized patterns.


\subsection{Limitations} 

The approach proposed in this article features several limitations. First, during the validation stage, only explorative experiments were performed. However, to get a better insight into which deep-learning models, augmentation pipelines and hyperparameters performed best a structural comparison is important. A structural comparison can for example include grid-search or ablation studies. Future work should include the investigation of more recent deep learning models, e.g., Residual Convolutional Neural Split-Attention Network (ResNeSt) \cite{ResNest}, Class-Attention in Image Transformers (CaiT) \cite{CaiT} or Data-efficient Image Transformers (DeiT) \cite{Deit}. This also applies to the experiments on the extended dataset. Due to time limitations during the validation stage of the challenge, no validation experiments were executed to identify the best-performing models for each class trained on the extended dataset. An attempt to clear the training dataset from augmented images using Scale-Invariant Feature Transform (SIFT) \cite{10.1023/b:visi.0000029664.99615.94} in order to achieve unbiased CV results was not successful. Future work should investigate further dataset cleansing strategies to address this problem. However, the exclusion of too similar images in the original datasets via hashing impedes the cleansing. More sophisticated ensembling strategies can further improve classification results.

Regarding the presented training dataset extension strategy both, the pseudo-labeling and synthetic image generation approach can be optimized. The threshold for inclusion of pseudo-label candidates was chosen conservatively on purpose and not evaluated via validation experiments. Hence, a more balanced choice is possible to achieve a greater or more qualitative outcome of additional training images. Consequently, these have a direct influence on the pix2pixHD models and extended EfficientNets model ensemble. For generated synthetic images via pix2pixHD models no metric-based analysis or quality assessment was conducted, hence there was no filtering of potentially harming samples. In addition, the visual assessment of these images was performed by non-clinicians. Hence, no statements on the actual convincibility regarding realistic looks in the eyes of clinicians can be made. Further, the applied class-balancing approach relying on massive amounts of synthetic images was rather aggressive. A more subtle approach with less extension for classes with an already sufficient amount of training images might enable better overall performance. In addition, unconditional GANs may have displayed a better choice for the given classification task as these do not require masks for training or generation. Respective recent developments such as StyleGAN2+ADA\footnote{StyleGAN2+ADA: \url{https://github.com/NVlabs/stylegan2-ada}, access 2021-09-22} \cite{karras2020stylegan2ada} further enable data efficiency via adaptive discriminator augmentation, facilitating qualitative results for rather small amounts of training images.


\section{Conclusion} 
\label{sec:conclusion}

This work investigated, whether training dataset extension with pseudo-labels and synthetic images generated by pix2pixHD can improve EfficientNet-based model ensemble performance for infection and ischaemia classification in DFUs. For evaluation, the amount of 5,955 labeled images of the training part of the DFUC 2021 dataset was extended with (i) 6,961 pseudo-labeled images from unlabeled images in the training and test part, and (ii) 38,748 synthetic images for subsequent class-balancing. The resulting extended training part had $8.67$ times the size of the baseline training dataset with a real to synthetic image ratio of $1:3$, featuring manifolds of synthetic images for the rare classes \texttt{ischaemia} and \texttt{both}.

Results show that the macro F1-Scores of the averaged baseline model ensembles outperformed the individual classifiers. All models trained on the extended dataset outperformed the baseline ensemble for the macro F1-Score. In particular, considerable improvements of the class F1-Scores for rare classes were achieved while no harming effects for common classes were detected. The best results were achieved for the averaged extended model ensemble.

Pseudo-labeling represents an effective strategy to extend datasets. Extension and class balancing via synthetic images generated by GANs has the potential to further improve the overall performance of classification models, especially that for rare classes, given a sufficient amount of images for training.


\section*{Acknowledgments}

Louise Bloch and Raphael Brüngel were partially funded by PhD grants from University of Applied Sciences and Arts Dortmund, Dortmund, Germany. The authors thank Henryk Birkh\"olzer for advice on pix2pixHD.



\bibliographystyle{spmpsci}
\bibliography{mybibliography}

\begin{thebibliography}{10}
\providecommand{\url}[1]{{#1}}
\providecommand{\urlprefix}{URL }
\expandafter\ifx\csname urlstyle\endcsname\relax
  \providecommand{\doi}[1]{DOI~\discretionary{}{}{}#1}\else
  \providecommand{\doi}{DOI~\discretionary{}{}{}\begingroup
  \urlstyle{rm}\Url}\fi

\bibitem{alzubaidi2019dfu_qutnet}
Alzubaidi, L., Fadhel, M.A., Oleiwi, S.R., Al-Shamma, O., Zhang, J.:
  {DFU\_QUTNet: Diabetic Foot Ulcer Classification using Novel Deep
  Convolutional Neural Network}.
\newblock Multimedia Tools and Applications \textbf{79}(21), 15,655--15,677
  (2019).
\newblock \doi{10.1007/s11042-019-07820-w}

\bibitem{Buslaev_2020}
Buslaev, A., Iglovikov, V.I., Khvedchenya, E., Parinov, A., Druzhinin, M.,
  Kalinin, A.A.: {Albumentations: Fast and Flexible Image Augmentations}.
\newblock Information \textbf{11}(2), 125 (2020).
\newblock \doi{10.3390/info11020125}

\bibitem{canny1986}
Canny, J.: {A Computational Approach to Edge Detection}.
\newblock {IEEE} Transactions on Pattern Analysis and Machine Intelligence
  \textbf{8}(6), 679--698 (1986).
\newblock \doi{10.1109/tpami.1986.4767851}

\bibitem{cassidy2021diabetic}
Cassidy, B., Kendrick, C., Reeves, N.D., Pappachan, J.M., O'Shea, C.,
  Armstrong, D.G., Yap, M.H.: Diabetic foot ulcer grand challenge 2021:
  Evaluation and summary.
\newblock arXiv preprint arXiv:2111.10376  (2021).
\newblock \urlprefix\url{https://arxiv.org/abs/2111.10376}

\bibitem{das2021dfu_spnet}
Das, S.K., Roy, P., Mishra, A.K.: {DFU\_SPNet: A Stacked Parallel Convolution
  Layers Based CNN to Improve Diabetic Foot Ulcer Classification}.
\newblock {ICT} Express  (2021).
\newblock \doi{10.1016/j.icte.2021.08.022}

\bibitem{das2021recognition}
Das, S.K., Roy, P., Mishra, A.K.: {Recognition of Ischaemia and Infection in
  Diabetic Foot Ulcer: A Deep Convolutional Neural Network Based Approach}.
\newblock International Journal of Imaging Systems and Technology  (2021).
\newblock \doi{10.1002/ima.22598}

\bibitem{Deng2009}
{Deng}, J., {Dong}, W., {Socher}, R., {Li}, L., Li, K., Fei-Fei, L.: {ImageNet:
  A Large-Scale Hierarchical Image Database}.
\newblock In: Proceedings of the {IEEE} Conference on Computer Vision and
  Pattern Recognition (CVPR 2009), pp. 248--255. IEEE (2009).
\newblock \doi{10.1109/cvpr.2009.5206848}

\bibitem{VisionTransformers}
Dosovitskiy, A., Beyer, L., Kolesnikov, A., Weissenborn, D., Zhai, X.,
  Unterthiner, T., Dehghani, M., Minderer, M., Heigold, G., Gelly, S.,
  Uszkoreit, J., Houlsby, N.: {An Image is Worth 16x16 Words: Transformers for
  Image Recognition at Scale}.
\newblock In: Proceedings of the 9th International Conference on Learning
  Representations (ICLR 2021). {ICLR} (2021)

\bibitem{falanga2005diabeticwoundhealing}
Falanga, V.: {Wound Healing and Its Impairment in the Diabetic Foot}.
\newblock The Lancet \textbf{366}(9498), 1736--1743 (2005).
\newblock \doi{10.1016/s0140-6736(05)67700-8}

\bibitem{galdran2021convolutional}
Galdran, A., Carneiro, G., Ballester, M.A.G.: Convolutional nets versus vision
  transformers for diabetic foot ulcer classification.
\newblock arXiv preprint arXiv:2111.06894  (2021).
\newblock \urlprefix\url{https://arxiv.org/abs/2111.06894}

\bibitem{goodfellow2014gans}
Goodfellow, I., Pouget-Abadie, J., Mirza, M., Xu, B., Warde-Farley, D., Ozair,
  S., Courville, A., Bengio, Y.: {Generative Adversarial Nets}.
\newblock In: Z.~Ghahramani, M.~Welling, C.~Cortes, N.~Lawrence, K.Q.
  Weinberger (eds.) Advances in Neural Information Processing Systems (NIPS
  2017), vol.~27. Curran Associates, Inc. (2014)

\bibitem{goyal2020dfunet}
Goyal, M., Reeves, N.D., Davison, A.K., Rajbhandari, S., Spragg, J., Yap, M.H.:
  {DFUNet: Convolutional Neural Networks for Diabetic Foot Ulcer
  Classification}.
\newblock {IEEE} Transactions on Emerging Topics in Computational Intelligence
  \textbf{4}(5), 728--739 (2020).
\newblock \doi{10.1109/tetci.2018.2866254}

\bibitem{ResNet}
He, K., Zhang, X., Ren, S., Sun, J.: {Deep Residual Learning for Image
  Recognition}.
\newblock In: Proceedings of the IEEE Conference on Computer Vision and Pattern
  Recognition (CVPR 2016), pp. 770--778 (2016).
\newblock \doi{10.1109/CVPR.2016.90}

\bibitem{Hinton2012RMSprop}
Hinton, G., Srivastava, N., Swersky, K.: {Lecture 6e rmsprop: Divide the
  Gradient by a Running Average of its Recent Magnitude} (2012).
\newblock
  \urlprefix\url{https://www.cs.toronto.edu/~tijmen/csc321/slides/lecture_slides_lec6.pdf}

\bibitem{isola2017}
Isola, P., Zhu, J.Y., Zhou, T., Efros, A.A.: {Image-to-Image Translation with
  Conditional Adversarial Networks}.
\newblock In: Proceedings of the {IEEE} Conference on Computer Vision and
  Pattern Recognition ({CVPR} 2017). {IEEE} (2017).
\newblock \doi{10.1109/cvpr.2017.632}

\bibitem{karras2020stylegan2ada}
Karras, T., Aittala, M., Hellsten, J., Laine, S., Lehtinen, J., Aila, T.:
  {Training Generative Adversarial Networks with Limited Data}.
\newblock In: H.~Larochelle, M.~Ranzato, R.~Hadsell, M.F. Balcan, H.~Lin (eds.)
  Advances in Neural Information Processing Systems (NeurIPS 2020), vol.~33,
  pp. 12,104--12,114. Curran Associates, Inc. (2020)

\bibitem{Krizhevsky2012}
Krizhevsky, A., Sutskever, I., Hinton, G.E.: {ImageNet Classification with Deep
  Convolutional Neural Networks}.
\newblock In: F.~Pereira, C.J.C. Burges, L.~Bottou, K.Q. Weinberger (eds.)
  Advances in Neural Information Processing Systems (NIPS 2012), vol.~25.
  Curran Associates, Inc. (2012)

\bibitem{10.1023/b:visi.0000029664.99615.94}
Lowe, D.G.: {Distinctive Image Features from Scale-Invariant Keypoints}.
\newblock International Journal of Computer Vision \textbf{60}(2), 91--110
  (2004).
\newblock \doi{10.1023/b:visi.0000029664.99615.94}

\bibitem{merkel2014docker}
Merkel, D.: {Docker: Lightweight Linux Containers for Consistent Development
  and Deployment}.
\newblock Linux journal \textbf{2014}(239), 2 (2014)

\bibitem{Micikevicius2018}
Micikevicius, P., Narang, S., Alben, J., Diamos, G., Elsen, E., Garcia, D.,
  Ginsburg, B., Houston, M., Kuchaiev, O., Venkatesh, G., Wu, H.: {Mixed
  Precision Training}.
\newblock In: Proceedings of the 6th International Conference on Learning
  Representations (ICLR 2018). {ICLR} (2018)

\bibitem{mirza2014cgans}
Mirza, M., Osindero, S.: {Conditional Generative Adversarial Nets}.
\newblock arXiv preprint arXiv:1411.1784  (2014).
\newblock \urlprefix\url{https://arxiv.org/abs/1411.1784}

\bibitem{PyTorch}
Paszke, A., Gross, S., Massa, F., Lerer, A., Bradbury, J., Chanan, G., Killeen,
  T., Lin, Z., Gimelshein, N., Antiga, L., Desmaison, A., Kopf, A., Yang, E.,
  DeVito, Z., Raison, M., Tejani, A., Chilamkurthy, S., Steiner, B., Fang, L.,
  Bai, J., Chintala, S.: {PyTorch: An Imperative Style, High-Performance Deep
  Learning Library}.
\newblock In: H.~Wallach, H.~Larochelle, A.~Beygelzimer, F.~d'Alch\'{e} Buc,
  E.~Fox, R.~Garnett (eds.) Advances in Neural Information Processing Systems
  (NeuriPS 2019), vol.~32, pp. 8024--8035. Curran Associates, Inc. (2019)

\bibitem{Ribeiro2016lime}
Ribeiro, M.T., Singh, S., Guestrin, C.: {Why Should I Trust You?: Explaining
  the Predictions of Any Classifier}.
\newblock In: Proceedings of the 22nd {ACM} International Conference on
  Knowledge Discovery and Data Mining (SIGKDD 2016), pp. 1135--1144 (2016).
\newblock \doi{10.1145/2939672.2939778}

\bibitem{Saeedi_2019}
Saeedi, P., Petersohn, I., Salpea, P., Malanda, B., Karuranga, S., Unwin, N.,
  Colagiuri, S., Guariguata, L., Motala, A.A., Ogurtsova, K., Shaw, J.E.,
  Bright, D., Williams, R.: {Global and Regional Diabetes Prevalence Estimates
  for 2019 and Projections for 2030 and 2045: Results from the International
  Diabetes Federation Diabetes Atlas, 9th edition}.
\newblock Diabetes Research and Clinical Practice \textbf{157}, 107,843 (2019).
\newblock \doi{10.1016/j.diabres.2019.107843}

\bibitem{sarp2021woundgan}
Sarp, S., Kuzlu, M., Wilson, E., Guler, O.: {WG2AN: Synthetic wound image
  generation using generative adversarial network}.
\newblock The Journal of Engineering \textbf{2021}(5), 286--294 (2021).
\newblock \doi{10.1049/tje2.12033}

\bibitem{siddiqui2010chronic_wound_infection}
Siddiqui, A.R., Bernstein, J.M.: {Chronic Wound Infection: Facts and
  Controversies}.
\newblock Clinics in Dermatology \textbf{28}(5), 519--526 (2010).
\newblock \doi{10.1016/j.clindermatol.2010.03.009}

\bibitem{EfficientNet}
Tan, M., Le, Q.: {EfficientNet: Rethinking Model Scaling for Convolutional
  Neural Networks}.
\newblock In: K.~Chaudhuri, R.~Salakhutdinov (eds.) Proceedings of the 36th
  International Conference on Machine Learning (ICML), \emph{Proceedings of
  Machine Learning Research (PMLR 2019)}, vol.~97, pp. 6105--6114. PMLR (2019)

\bibitem{EfficientNetv2}
Tan, M., Le, Q.: {{E}fficient{N}et: Rethinking Model Scaling for Convolutional
  Neural Networks}.
\newblock In: K.~Chaudhuri, R.~Salakhutdinov (eds.) Proceedings of the 36th
  International Conference on Machine Learning (ICML), \emph{Proceedings of
  Machine Learning Research (PLMR 2019)}, vol.~97, pp. 6105--6114. PMLR (2019)

\bibitem{Deit}
Touvron, H., Cord, M., Douze, M., Massa, F., Sablayrolles, A., Jegou, H.:
  {Training Data-Efficient Image Transformers Distillation Through Attention}.
\newblock In: Proceedings of the International Conference on Machine Learning
  (ICML 2021), vol. 139, pp. 10,347--10,357 (2021)

\bibitem{CaiT}
Touvron, H., Cord, M., Sablayrolles, A., Synnaeve, G., J\'egou, H.: {Going
  Deeper With Image Transformers}.
\newblock In: Proceedings of the IEEE/CVF International Conference on Computer
  Vision (ICCV 2021), pp. 32--42 (2021)

\bibitem{wang2018}
Wang, T.C., Liu, M.Y., Zhu, J.Y., Tao, A., Kautz, J., Catanzaro, B.:
  {High-Resolution Image Synthesis and Semantic Manipulation with Conditional
  GANs}.
\newblock In: Proceedings of the {IEEE}/{CVF} Conference on Computer Vision and
  Pattern Recognition ({CVPR} 2018). {IEEE} (2018).
\newblock \doi{10.1109/cvpr.2018.00917}

\bibitem{rw2019timm}
Wightman, R.: {PyTorch Image Models}.
\newblock \url{https://github.com/rwightman/pytorch-image-models} (2019).
\newblock \doi{10.5281/zenodo.4414861}

\bibitem{10.1109/BHI50953.2021.9508563}
Yap, M.H., Cassidy, B., Pappachan, J.M., O’Shea, C., Gillespie, D., Reeves,
  N.D.: {Analysis Towards Classification of Infection and Ischaemia of Diabetic
  Foot Ulcers}.
\newblock In: Proceedings of the IEEE EMBS International Conference on
  Biomedical and Health Informatics (BHI 2021), pp. 1--4 (2021).
\newblock \doi{10.1109/BHI50953.2021.9508563}

\bibitem{yap2021detection}
Yap, M.H., Hachiuma, R., Alavi, A., Brüngel, R., Cassidy, B., Goyal, M., Zhu,
  H., Rückert, J., Olshansky, M., Huang, X., Saito, H., Hassanpour, S.,
  Friedrich, C.M., Ascher, D.B., Song, A., Kajita, H., Gillespie, D., Reeves,
  N.D., Pappachan, J.M., O{\textquotesingle}Shea, C., Frank, E.: {Deep Learning
  in Diabetic Foot Ulcers Detection: A Comprehensive Evaluation}.
\newblock Computers in Biology and Medicine \textbf{135}, 104,596 (2021).
\newblock \doi{10.1016/j.compbiomed.2021.104596}

\bibitem{dfuc2020}
Yap, M.H., Reeves, N., Boulton, A., Rajbhandari, S., Armstrong, D., Maiya,
  A.G., Najafi, B., Frank, E., Wu, J.: {Diabetic Foot Ulcers Grand Challenge
  2020}.
\newblock \doi{10.5281/zenodo.3731068}

\bibitem{dfuc2021}
Yap, M.H., Reeves, N., Boulton, A., Rajbhandari, S., Armstrong, D., Maiya,
  A.G., Najafi, B., Frank, E., Wu, J.: {Diabetic Foot Ulcers Grand Challenge
  2021}.
\newblock \doi{10.5281/zenodo.3715020}

\bibitem{dfuc2022}
Yap, M.H., Reeves, N., Boulton, A., Rajbhandari, S., Armstrong, D., Maiya,
  A.G., Najafi, B., Frank, E., Wu, J.: {Diabetic Foot Ulcers Grand Challenge
  2022}.
\newblock \doi{10.5281/zenodo.4575228}

\bibitem{ResNest}
Zhang, H., Wu, C., Zhang, Z., Zhu, Y., Lin, H., Zhang, Z., Sun, Y., He, T.,
  Mueller, J., Manmatha, R., Li, M., Smola, A.: {ResNeSt: Split-Attention
  Networks}.
\newblock arXiv preprint arXiv:2004.08955  (2020).
\newblock \urlprefix\url{https://arxiv.org/abs/2004.08955}

\bibitem{zhang2018woundgan}
Zhang, J., Zhu, E., Guo, X., Chen, H., Yin, J.: {Chronic Wounds Image Generator
  Based on Deep Convolutional Generative Adversarial Networks}.
\newblock In: Communications in Computer and Information Science, pp. 150--158.
  Springer Singapore (2018).
\newblock \doi{10.1007/978-981-13-2712-4\_11}

\bibitem{zhang2016dfu_prevalence}
Zhang, P., Lu, J., Jing, Y., Tang, S., Zhu, D., Bi, Y.: {Global Epidemiology of
  Diabetic Foot Ulceration: A Systematic Review and Meta-Analysis}.
\newblock Annals of Medicine \textbf{49}(2), 106--116 (2016).
\newblock \doi{10.1080/07853890.2016.1231932}

\end{thebibliography}
\end{document}